\documentclass{article}

\usepackage{microtype}
\usepackage{graphicx}
\usepackage{subcaption}
\usepackage{booktabs} 
\usepackage{multirow}
\usepackage{multicol}
\usepackage{hyperref}
\usepackage{enumitem}

\usepackage{algorithmic, algorithm}
\usepackage[preprint]{icml2026}

\usepackage{amsmath}
\usepackage{amssymb}
\usepackage{mathtools}
\usepackage{amsthm}

\usepackage[capitalize,noabbrev]{cleveref}

\theoremstyle{plain}

\theoremstyle{definition}

\theoremstyle{remark}

\usepackage[textsize=tiny]{todonotes}

\renewcommand{\algorithmiccomment}[1]{\hfill$\triangleright$ #1}

\icmltitlerunning{Retrieve-Verify-Retrieve}
\begin{document}

\twocolumn[
  \icmltitle{RVR: Retrieve-Verify-Retrieve for Comprehensive Question Answering}



  \icmlsetsymbol{equal}{*}

  \begin{icmlauthorlist}
    \icmlauthor{Deniz Qian}{equal,yyy}
    \icmlauthor{Hung-Ting Chen}{equal,yyy}
    \icmlauthor{Eunsol Choi}{yyy}
  \end{icmlauthorlist}

  \icmlaffiliation{yyy}{Courant Institute School of Mathematics, Computing, and Data Science, New York University}

  \icmlcorrespondingauthor{Deniz Qian}{dq2024@nyu.edu}
  \icmlcorrespondingauthor{Hung-Ting Chen}{hc3337@nyu.edu}

  \icmlkeywords{Machine Learning, ICML}

  \vskip 0.3in
]



\printAffiliationsAndNotice{\icmlEqualContribution}

\begin{abstract}
  Comprehensively retrieving diverse documents is crucial to address queries that admit a wide range of valid answers. We introduce retrieve-verify-retrieve (RVR), a multi-round retrieval framework designed to maximize answer coverage. Initially, a retriever takes the original query and returns a candidate document set, followed by a verifier that identifies a high-quality subset. For subsequent rounds, the query is augmented with previously verified documents to uncover answers that are not yet covered in previous rounds. RVR is effective even with off-the-shelf retrievers, and fine-tuning retrievers for our inference procedure brings further gains. Our method outperforms baselines, including agentic search approaches, achieving at least 10\% relative and 3\% absolute gain in complete recall percentage on a multi-answer retrieval dataset (QAMPARI). We also see consistent gains on two out-of-domain datasets (QUEST and WebQuestionsSP) across different base retrievers. Our work presents a promising iterative approach for comprehensive answer recall leveraging a verifier and adapting retrievers to a new inference scenario.  
  
  
  


\end{abstract}

\section{Introduction}
\label{introduction}
Retrieval is a key method to equip large language models (LLMs) with up-to-date, long-tail information. Despite the recent advances in retrieval systems, comprehensively recovering relevant documents remains challenging~\cite{amouyal2022qampari,chen-choi-2025-open}.   



\begin{figure}[t]
\begin{center}
\centerline{\includegraphics[width=1.0\columnwidth]{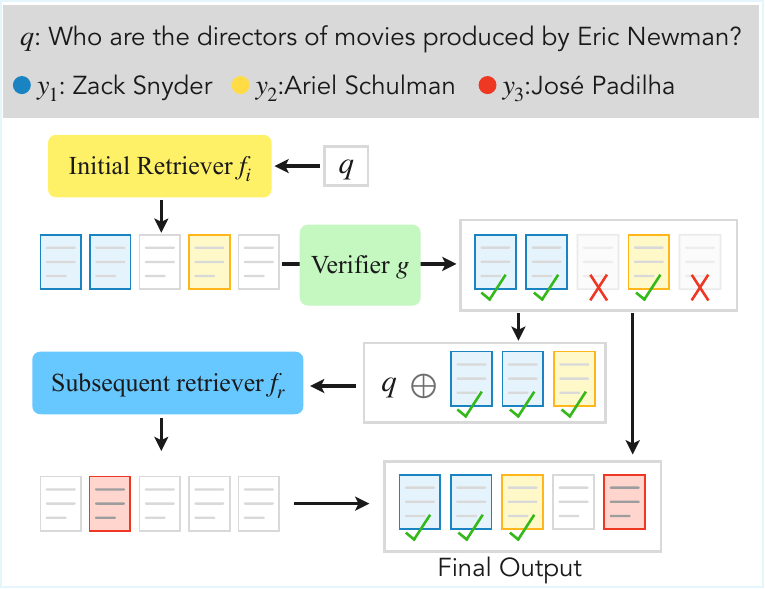}}
\caption{Overview of our Retrieve-Verify-Retrieve framework. Each query $q$ aims to retrieve documents to get multiple answers $(y_1, y_2,y_3)$. The initial retriever takes a query and returns document sets, and the verifier examines each document, identifying two valid answers $y_1, y_2$. The subsequent retriever takes the query and documents containing identified answers as input, targeting to retrieve complementary answer $y_3$.}
\label{fig:intro-figure}
\end{center}
\vskip -0.2in
\end{figure}

In this work, we introduce Retrieve-Verify-Retrieve (RVR), a framework that performs multiple rounds of retrieval, where each round conditions on previously retrieved documents verified by an LLM. At the next iteration, the model receives the concatenation of the query and LLM verified output documents from the previous retrieval search, aiming to search for remaining gold documents. The high-quality subsets from the initial step and the outputs returned by the subsequent retriever are merged as the final output. Figure~\ref{fig:intro-figure} visualizes our approach. 

Unlike traditional retrievers that only take a query as input to retrieve documents, our retriever is trained to explicitly condition on previously retrieved, high-quality documents to target missing ones. This formulation encourages improving coverage over iterations, in contrast to a one-shot ranking task between the original query and documents. Recent agentic approaches~\cite{jin2025search, tongyidr} also allows iterative search by interweaving retrieval calls and new query formulation, but they mainly targets multi-hop questions which require a sequence of distinct search queries, not multiple retrieval rounds targeting comprehensive answers for the same query. 






Our approach is trained and evaluated on an open-domain, multi-answer QA benchmark QAMPARI \citep{amouyal2022qampari}. Compared to base fine-tuned retrievers as well as recent agentic search framework~\cite{jin2025search, tongyidr}, our approach shows consistent improvements in retrieval performance. Using off-the-shelf retrievers iteratively also showed some gains, but larger gains are achieved by fine-tuning retrievers with our inference setting. We further evaluate on other multi-answer QA benchmarks QUEST ~\cite{malaviya2023quest} and WebQuestionsSP ~\cite{yih-etal-2016-value}, showing that our approach can generalize. To understand the impact of verifier performance, we provide performance under both our oracle verifier and our verifier. The experimental results show oracle verifier can further significantly boost the performance, showing the headroom of the approach with better verifier models. 

In summary, we present a iterative retrieval framework for comprehensive retrieval. Despite rich studies on agentic search, most approaches focus on improving LLMs that generate new queries based on retrieval output rather than adapting retrievers for new inference scenarios. Our results support fine-tuning retrievers for new inference scenarios can bring further performance gains. We release the code and model at \url{https://github.com/timchen0618/RVR}.



\section{Related Work}
\label{related work}
\paragraph{Comprehensive Retrieval}
We review prior retrieval datasets with queries that admit a list of answers. QAMPARI~\citep{amouyal2022qampari} and QUEST~\citep{malaviya2023quest} expect a list of entity answers from Wikipedia; \citet{zhu-etal-2024-fanoutqa} introduces FanOutQA, which requires multi-hop reasoning over large numbers of documents to aggregate information about multiple entities; ~\citet{katz-etal-2023-neretrieve} expects all documents mentioning a list of entities given an entity type; \citet{yih-etal-2016-value} provide WebQuestionsSP, which annotates questions with SPARQL semantic parses over Freebase for knowledge base question answering; ~\citet{chen-choi-2025-open} expect the candidate document set to cover multiple valid perspectives given a subjective question.

In terms of method, \citet{min2021jointpr} introduces an autoregressive reranker conditioned on previously selected passages to encourage diversity and cover multiple answers. ~\citet{chen2025beyond} proposed to generate multiple query embeddings autoregressively to retrieve more comprehensive document sets. In contrast to both, we chose iterative framework, where the output from previous retrieval step is provided as an input to the next retrieval stage, similar to recent agentic approaches. 

\paragraph{Iterative \& Agentic Retrieval}
Several works have explored performing multiple rounds of retrieval for complex question answering~\citep{yang2018hotpotqa,trivedi2022musique}. \citet{qi2019iterativequery} introduced iterative query reformulation, generating new retrieval queries from partially read passages. A line of work~\cite{xiong2020multihop,trivedi2023ircot,fang2025kiragknowledgedriveniterativeretriever} similarly retrieve evidence iteratively, interweaving with intermediate LLM reasoning steps. A more recent line of works on agentic search systems~\citep{jin2025search,tongyidr,shao2025dr} explored using retrievers as a tool, and an LLM agent alternates between reasoning and tool calling until it reaches the final answer. 
Concurrent work~\citep{sharifymoghaddam2026rerank} shows that retrieving more documents at each turn and reranking them can improve answer accuracy. While most agentic approaches use off-the-shelf retrievers, \citet{liu2026agentic} train retrievers for agentic search by optimizing for both local relevance and global answer correctness in single-answer reasoning tasks.

\paragraph{Verifier Based Retrieval}
Recent work has incorporated verification into retrieval pipelines. Chain-of-Verification (CoVe) \citep{dhuliawala2024chainofverification} has an LLM draft an answer, plan verification questions, retrieve supporting evidence, and revise, reducing hallucinations. Self-RAG~\citep{asai2024selfrag} integrates retrieval, generation, and critique via self-reflection signals. In these approaches, verification operates \emph{after} retrieval or generation, serving primarily as a filtering mechanism. In contrast, our framework integrates verification \emph{within} the retrieval loop itself: verification determines which retrieved documents are retained and used to condition subsequent retrieval rounds. 

\section{Method}
\label{method}

\subsection{Problem Formulation} 
\paragraph{Multi Answer Retrieval Task}
Given a query $q$ and a large corpus $\mathcal{C} = \{d_1, \dots, d_N\}$, 
a retrieval system $f(q, \mathcal{C}, K)$ should identify a ranked subset of $K$ documents 
$D_\text{out} = \{d_1, \dots, d_K\} \subset \mathcal{C}$ that contains all answers $Y={y_1, y_2,...,y_M}$ for the query $q$. A retriever should aim to maximize both \emph{relevance} and \emph{coverage}, 
ensuring that the retrieved set reflects the full range of relevant, high-quality information.

\paragraph{Metrics} We assume a test set, where each query $q$ is annotated with a set of $M$ distinct gold answer strings $Y={y_1, y_2,...,y_M}$. We judge that document $d_i$ covers answer $y_j$ if the substring occurs in the document. We report the following two metrics related to task performance:
\begin{itemize}[noitemsep,leftmargin=10px]
    \item \textbf{MRecall@K}: a binary score that 
    equals 1 if all answers or at least $K$ answers in the answer set $\{y_1...y_M\}$ are covered by $D_\text{out}$. This metric was introduced in prior work~\cite{min2021jointpr} for questions that admit multiple valid answers.
    \item \textbf{Recall@K}: the fraction of answers $Y$ that are covered by at least one document in $D_\text{out}$. This is a less stringent metric compared to MRecall@K. 
\end{itemize}

\begin{algorithm}

\caption{Inference procedure for Retrieve-Verify-Retrieve framework.}

\label{alg:inference}
\centering
\vspace{0.5em}
\begin{minipage}{0.47\textwidth}

\textbf{Input:} A input query $q$, corpus $\mathcal{C}$, initial retriever $f_i$, subsequent retriever $f_r$, a verifier $g$.\\[0.2em]
\textbf{Hyperparameters:} verifier budget per turn $B$, max document context budget $M$, maximum number of turns $T$, final output set size $K$.\\[0.2em]
\textbf{Output:} A ranked set of documents $D_{\text{out}} = \{d_1, \ldots, d_K\} \subset \mathcal{C}$, sorted by relevance to $q$.
\end{minipage}

\vspace{1em}
\small
\begin{algorithmic}[1]
\STATE $D_{\text{out}} \gets \emptyset$
\STATE $D_i \gets f_i(q, \mathcal{C}, K)$
\FOR{$t \in \{1, \ldots, T\}$ \textbf{while} $|D_{\text{out}}| < K$}
    \STATE $D_{\text{v}} \gets \{d \in D_i : g(d, q) \land \operatorname{rank}(d) \leq B \}$ \COMMENT{verified subset}
    \STATE $D_{\text{out}} \gets D_{\text{out}} \cup D_{\text{v}}$
    \STATE $D_{\text{ctx}} \gets \textsc{TopK}(M, D_{\text{v}})$ \COMMENT{first $M$ for context}
    \STATE $q_r \gets [q ; \bigoplus_{d \in D_{\text{ctx}}} d]$ \COMMENT{concatenate query with context}
    \STATE $D_i \gets f_r(q_r, \mathcal{C}, K)$ \COMMENT{perform $t+1$-th retrieval}
\ENDFOR
\STATE $D_{\text{out}} \gets D_{\text{out}} \cup D_i$ \COMMENT{add remaining}
\STATE \textbf{return} $\textsc{TopK}(K, D_{\text{out}})$
\end{algorithmic}
\small
\hrulefill
\vspace{-0.5em}

\begin{itemize}
    \setlength{\itemsep}{0pt}
    \setlength{\parskip}{0pt}
    \item $\{d \in D : P(d)\}$: elements in $D$ satisfying predicate $P$
    \item $\textsc{TopK}(k, D)$: first $k$ elements of ordered set $D$
    \item $\bigoplus_{d \in D} d$: concatenation over ordered set $D$
    \item $\operatorname{rank}(d)$: position of document $d$ in the retrieval ranking
\end{itemize}
\vspace{-1em}
\label{fig:inference-formula}
\end{algorithm}

\subsection{Iterative Retrieval Framework} \label{sec:iter-framework}

Algorithm~\ref{alg:inference} describes our iterative retrieval system. Our framework uses three components: an initial retriever $f_i$ that uses only the query, an iterative retriever $f_r$ that conditions on both the query and previously retrieved documents, and a verifier model $g$. Our subsequent retriever $f_r$ can be viewed as a version of task-aware retrieval with instruction~\cite{Asai2022TaskawareRW} which is trained to retrieve documents closer to the query $q$ but distinct from documents in the input ($\bigoplus_{d \in D_{\text{ctx}}} d$). The initial and subsequent retrievers can be the same or different retrieval models, and we evaluate both possibilities in our experiments. 

\paragraph{Initial Retriever}
The initial retriever produces a ranked list of $K$ documents ($D_i = f_i(q, \mathcal{C}, K)$, line 2).

\paragraph{Verifier}
Retrieval is often noisy, and contains irrelevant documents.
Thus, we implement a verifier $g$ that examines retrieval outputs and produces a verified subset $D_V = \{d \in D_i : g(d, q) \wedge \text{rank}(d) \leq B\}$ containing documents deemed relevant by verifier $g$, where $B$ is the verifier budget (line 4).

The verified subset, $D_V$, will constitute the final output (line 10) and will be used in forming a query for the next stage retrieval. For subsequent retrieval turns, we select the top $M$ documents from the verified set (line 6) to form the concatenated query $q_r = [q; \bigoplus_{d \in D_{\text{ctx}}} d]$. 

\paragraph{Subsequent Retriever}
Using this query, the subsequent retriever produces new output $
D_i = f_r(q_r, \mathcal{C}, B)$ (line 8). Using augmented query $q_r$ allows $f_r$ to reason about which relevant documents remain unretrieved, promoting answer coverage across iterations.

\paragraph{Output} To form the final output set, we accumulate verified documents across all rounds and add any remaining documents from the final iteration. Set semantics remove duplicates $
D_{\text{out}} = \operatorname{TopK}(K, \Big( \bigcup_{t=0}^{T-1} D_V^{(t)} \Big) \;\cup\;D_{i}^T)$ where $D_V^{(t)}$ denotes the verified set at round $t$, and we take only the top $K - |D_{\text{out}}|$ documents from the final retrieval $D_i^{(T)}$ to ensure the total output size is $K$.

\subsection{Training} \label{subsec:train}

In this section, we describe the training of two sets of retrievers $f_i$ and $f_r$. We assume a training data, where each instance contains a query $q$ paired with a set of gold documents $D^*$ from corpus $\mathcal{C}$. We do not train the verifier, and use off-the-shelf LLMs. Our novelty lies in training the retriever $f_r$ with the proposed iterative inference scenario. We thus generate training data (positive and negative target documents) from this inference scenario. Besides that, we use standard contrastive retriever learning objective~\citep{izacard2021unsupervised}:
\[
\mathcal{L_\theta} =
-\log \frac{\exp(s(f_\theta(x), f_\theta(d^{+}))/\tau)}
{\sum_{d^ \in D_{batch}}\exp(s(f_\theta(x), f_\theta(d))/\tau)}
\]
where retriever $f_\theta$ encodes the input query $x$, positive document $d^+$, and in-batch documents $d$, $s$ is a similarity function between the two embeddings, and $\tau$ is a temperature hyperparameter. We describe the process of constructing training data $(x, d^+, D_{batch})$ below.

\paragraph{Training Data for Initial Retriever ($D_i$)}
Each query $q$ is paired with one positive document $d^+$ and a set of negative documents $D_{-}$.
We also randomly sample one negative document $d^-$ from the corpus $\mathcal{C}$. Additionally, we leverage in-batch negatives~\cite{dpr} such that all the other positive and negative documents from other training examples in the same batch serve as additional negatives. Thus, $D_{batch}$ denotes all the documents in the batch, including the positive document, the sampled negative, and the in-batch negatives. We use the input query $q$ as is as $x$, forming ($x, d^+, D_{batch}$).

\paragraph{Training Data for Subsequent Retriever ($D_r$)}
As discussed in Section~\ref{sec:iter-framework}, $f_r$ also takes gold documents as input in addition to query $q$, where the number of appended gold documents is controlled by hyperparameter $M$. We first uniformly sample an integer $m \in \{0, ..., \min (M, |D^*|)\}$, and then sample $m$ documents from our ground truth set $D^*$ to form context $D_{\text{ctx}}$. The input query $x$ will be $[q; D_{\text{ctx}}]$. The positive document $d^+$ is randomly selected from the $D^*\setminus D_{\text{ctx}}$. Same as the training data for initial retriever, we randomly sample one negative document $d^-$ from the corpus $\mathcal{C}$, and leverage in-batch negatives.

\subsection{Implementation Details}

\paragraph{Retriever} 
For all experiments, we initialize the retrievers with pre-trained, off-the-shelf dual encoder retrievers: Contriever-MSMARCO~\citep{izacard2021unsupervised}, Qwen3-Embedding-0.6B \citep{qwen3embedding}, and INF-Retriever-v1-1.5B \citep{infly-ai_2025}.

We use instances from the training portion of the QAMPARI dataset. Every model is fine-tuned for $50$k steps. We use a learning rate of \(1\times10^{-4}\), temperature \(\tau=0.05\), batch size 48 per GPU, gradient accumulation of 2 steps, warm-up for 1K steps, and the AdamW optimizer \citep{loshchilov2018decoupled}. We use in-batch negatives and sample one negative document from the corpus. All experiments are conducted on NVIDIA H200 and L40 GPUs.

\paragraph{Verifier}
We prompt an LLM (Qwen \citep{qwen3technicalreport})\footnote{\texttt{Qwen/Qwen3-30B-A3B-Instruct-2507}} to serve as a verifier. Given an input document $d$ and a query $q$, the verifier outputs a binary label whether $d$ is relevant to the query $q$. Formally, $g(d,q) = \{0, 1\}$. We use vLLM~\citep{kwon2023efficient} for inference. See Appendix~\ref{appendix:qwen3} for the prompt, and we discuss verifier performance in Section~\ref{subsec:verifier_performance}.

\begin{table*}[t]
\caption{Main Experimental Results on QAMPARI test set (N=1000). We report MRecall@100 and Recall@100. Our proposed methods outperform both the base retrievers (Base) and retrievers fine-tuned in-domain (FT ($D_i$)). RVR approaches also achieve much higher performance than agentic baselines. For our results, statistical significance is tested using bootstrap resampling with 10,000 trials at $\alpha = 0.05$. $\dagger$ indicates statistically significant improvement over FT$(D_i)$. * indicates statistically significant improvement over FT$(D_i)$ + FT$(D_i)$.}
\label{tab:qampari-abs}
\vskip 0.15in
\begin{center}
\begin{small}
\setlength{\tabcolsep}{4pt}
\renewcommand{\arraystretch}{0.98}
\begin{tabular}{lcc|cc|cc}
\toprule
& \multicolumn{6}{c}{\textbf{Base Retriever Model}}\\
& \multicolumn{2}{c}{\textbf{Contriever-MSMARCO}} &
\multicolumn{2}{|c|}{\textbf{Qwen3-Embedding-0.6B}} &
\multicolumn{2}{c}{\textbf{INF-Retriever-v1-1.5B}} \\
\cmidrule(lr){2-3} \cmidrule(lr){4-5} \cmidrule(lr){6-7}
& MR@100 & R@100 & MR@100 & R@100 & MR@100 & R@100 \\ \midrule
 Base $f_i$ & 19.00 & 54.17 & 16.70 & 52.94 & 26.10 & 62.34\\
 FT $(D_i)$ & 28.60 & 63.19 & 26.90 & 63.48 & 29.30 & 65.99\\
\midrule
Tongyi (w/ Base $f_i$) & 6.60 & 35.96 & 16.20 & 52.22 & 20.40 & 57.03\\ 
Tongyi (w/ FT $(D_i)$) & 9.80 & 42.09 & 22.00 & 60.67 & 21.30 & 58.13\\
SearchR1 (w/ Base  $f_i$) & 7.40 & 36.58 & 17.60 & 53.39 & 21.40 & 57.72\\
SearchR1 (w/ FT $(D_i)$) & 9.60 & 40.83 & 24.30 & 57.82 & 27.00 & 60.96\\ \midrule
\multicolumn{7}{l}{\textbf{Ours: Retrieve-Verify-Retrieve}}\\
\hspace{1em} FT $(D_i)$ + FT $(D_i)$ & 28.80 & 63.59 & 30.30$^{\dagger}$ & 66.80$^{\dagger}$ & 31.10$^{\dagger}$ & 66.76$^\dagger$ \\ 
\hspace{1em} FT ($D_i$ + $D_r$) +  FT ($D_i$ + $D_r$) & \textbf{31.70}$^{\dagger*}$ & 66.12$^{\dagger*}$ & 29.20$^{\dagger}$ & 65.73$^{\dagger}$ & 32.40$^{\dagger}$ & 68.04$^{\dagger}$ \\ 
\hspace{1em}   FT $(D_i)$ + FT $(D_r)$     & {31.60}$^{\dagger*}$ & \textbf{66.83$^{\dagger*}$} & \textbf{31.40}$^{\dagger*}$ & \textbf{67.28}$^{\dagger*}$ & \textbf{33.70}$^{\dagger*}$ & \textbf{68.70}$^{\dagger*}$\\\bottomrule
 \end{tabular}
 \end{small}
 \end{center}
 \end{table*}

\begin{table*}[t]
\caption{Efficiency comparison. We report the number of retrieval calls (\# Calls) per query, as well as the seconds taken for retrieval (Ret.), the seconds taken for verification (Ver.) and total (sum of Ret. and Ver.). RVR methods are significantly more efficient than agentic baselines but still achieve better performance. }
\label{tab:efficiency_time}
\vskip 0.15in
\begin{center}
\begin{small}
\setlength{\tabcolsep}{4pt}
\renewcommand{\arraystretch}{0.98}
\begin{tabular}{l cccc| cccc| cccc}
\toprule
& \multicolumn{4}{c|}{\textbf{Contriever-MSMARCO}} & \multicolumn{4}{c|}{\textbf{Qwen3-Embedding-0.6B}} & \multicolumn{4}{c}{\textbf{INF-Retriever-v1-1.5B}} \\
\cmidrule(lr){2-5} \cmidrule(lr){6-9} \cmidrule(lr){10-13}
& {\# Calls $\downarrow$} & Ret.$\downarrow$ & Ver.$\downarrow$ & Sum$\downarrow$ & {\# Calls$\downarrow$} & Ret.$\downarrow$ & Ver.$\downarrow$ & Sum$\downarrow$ & {\# Calls $\downarrow$} & Ret.$\downarrow$ & Ver.$\downarrow$ & Sum $\downarrow$ \\
\midrule
Base & \textbf{1} & \textbf{1.91} & - & \textbf{1.91} & \textbf{1} & \textbf{1.34} & - & \textbf{1.34} & \textbf{1} & \textbf{3.75} & - & \textbf{3.75} \\
Tongyi &14.94&97.72&92.32&190.04&15.85&232.79&83.60&316.39&16.42&264.30&80.40&344.70\\
SearchR1 &2.92&19.43&2.41&21.84&2.96&22.95&2.42&25.38&2.93&39.86&2.54&42.40 \\
RVR (w/ T=2, B=1) & 2& 3.80 & 0.07 & 3.87 & 2& 5.05& 0.06 & 5.11 &2 & 7.55 & 0.07 & 7.62\\
RVR (w/ T=2, B=50) &2 & 3.80& 0.53 & 4.33 &2& 5.05& 0.52 &5.57 &2 & 7.55 & 0.53 & 8.08\\
RVR (w/ T=2, B=100) & 2& 3.80& 0.98 & 4.78 &2 & 5.05 &0.97 & 6.02&2 & 7.55 & 0.97 & 8.52 \\
\bottomrule
\end{tabular}
\end{small}
\end{center}
\vskip -0.1in
\end{table*}

\begin{table}[t]
\caption{Memory requirement comparison (all in GB), for LLM used $\downarrow$ and retriever index $\downarrow$. We report the size of each retriever model next to its name in the first row. }
\label{tab:efficiency_memory}
\vspace{-0.4em}

\begin{center}
\begin{small}
\setlength{\tabcolsep}{3pt}
\renewcommand{\arraystretch}{0.98}
\begin{tabular}{l cc| cc| cc}
\toprule
& \multicolumn{2}{c|}{\textbf{Contr. (0.48)}} & \multicolumn{2}{c|}{\textbf{Qwen3 (1.50)}} & \multicolumn{2}{c}{\textbf{INF (5.43)}} \\
 \cmidrule(lr){2-3} \cmidrule(lr){4-5} \cmidrule(lr){6-7}
& LLM  &Index &  LLM  & Index & LLM  & Index \\
\midrule
Base & - & 73.98 & - & 98.63 & -& 147.96\\
Tongyi & 61.06&74.30 & 61.06& 99.35& 61.06& 139.01 \\
SearchR1 & 17.41 &73.98 & 17.41& 98.63 & 17.41&147.95 \\
RVR ($f_i$==$f_r$) & 56.93  & 73.98 & 56.93  & 98.63 &56.93 & 147.96 \\
RVR ($f_i$!=$f_r$)
& 56.93  & 147.96 & 56.93  & 197.26 &56.93 & 295.92 \\
\bottomrule
\end{tabular}
\end{small}
\end{center}
\vskip -0.1in
\end{table}
\vspace{-0.2em}
\section{Experimental Settings}
\label{experiments}
\subsection{Dataset} 
\paragraph{QAMPARI} consists of open-domain questions paired with multiple valid answer strings (one example instance is shown in Fig~\ref{fig:intro-figure}). {On average, each query is annotated with a set of $16.58$ gold documents and contains $14.43$ unique valid answers.} Unlike most other datasets such as HotpotQA~\citep{yang2018hotpotqa} and BrowseCompPlus~\citep{chen2025browsecompplusfairtransparentevaluation} which assume one gold answer, this dataset allows studying multi-answer coverage. 
We use their original split (training, evaluation). 


\paragraph{QUEST} consists of queries paired with multiple valid answers, where each answer corresponds to a relevant Wikipedia document. The queries in QUEST specify set operations such as intersection, union, and difference (e.g., ``what are some Films about bats that are not Superhero films" (difference), ``films that are South Korean adventure comedies or Canadian fantasy comedies" (union)). The answer set size is 10.5 per query on average.

\paragraph{WebQuestionsSP} consists of questions from the Google Suggest API with their semantic parses and corresponding answer entities in Freebase. We treat WebQuestionsSP as a multi-answer retrieval task as there are on average $8.75$ answers per query.

\paragraph{Retrieval Corpus} For all datasets, we follow~\citet{amouyal2022qampari} and use a Wikipedia dump from 2021-08-01 consisting of 25.9 million passages, averaging 100 words. See Appendix~\ref{appendix:qampari-stat}, ~\ref{appendix:quest-stat}, and ~\ref{appendix:webqsp-stat} for full data statistics.





\subsection{Comparison Systems}

\paragraph{Base $f_i$}
We evaluate three pre-trained off-the-shelf retrieval models without fine-tuning.
\paragraph{FT ($D_i$)} Each pre-trained retriever is fine-tuned on the QAMPARI training dataset with standard contrastive objective (initial retriever setting in Section~\ref{subsec:train}). 



 \paragraph{Agentic Retrievers}
 We implement two strong open-source agentic baselines, Tongyi DeepResearch agent\footnote{\texttt{Alibaba-NLP/Tongyi-DeepResearch-30B-A3B}}~\cite{tongyidr} and SearchR1\footnote{\texttt{PeterJinGo/SearchR1-nq\_hotpotqa\_train-} \texttt{qwen2.5-7b-em-ppo}}~\cite{jin2025search}. Both approaches use fixed retriever and train a LLM that generates search query. The former is continued pre-trained on agentic trajectories and post-trained on synthetic QA pairs. The latter is trained using PPO~\citep{schulman2017proximal} on Natural Questions~\citep{kwiatkowski-etal-2019-natural} and HotpotQA to maximize answer accuracy.

 We use their trained model as is and follow their original implementations to retrieve $k_t$ = 5 documents for Tongyi and $k_s$ = 3 documents for SearchR1, respectively. To obtain the final candidate document set, we combine the documents returned for each retriever call without duplicates. For each agent, we run until it outputs an answer or it has collected over $K$ = 100 documents in total. If the agent has only collected $K_a$ documents ($K_a < K$) upon termination, we additionally retrieve a set of $(K - K_a)$ documents using the last query issued to the retriever and append them to the candidate document list. 
 
 The exact prompts we use are provided in Appendix~\ref{appendix:searchr1-prompt} and~\ref{appendix:tongyi-prompt}, and example trajectory traces are provided in Appendix~\ref{appendix:searchr1-traj} and~\ref{appendix:tongyi-traj}. We experiment them with two retriever configurations, base retriever and fine-tuned retriever for fair comparison. 

\paragraph{Retrieve-Verify-Retrieve} 
We use the following hyperparameters for our approach: $T$ = 2 retrieval rounds, verifier budget $B$ = 100, context budget $M$ = 3 for Contriever-MSMARCO and $M$ = 6 for INF-Retriever and Qwen3-Embedding. We evaluate three RVR configurations, differing in what is used for initial retriever ($f_i$) and subsequent retriever ($f_r$) : (1) FT ($D_i$) + FT ($D_i$): uses the same fine-tuned initial retriever $f_i$ in both rounds; (2) FT ($D_i$ + $D_r$) + FT ($D_i$ + $D_r$): uses a single model trained on the union of two retrieval training data ($D_i$ and $D_r$); (3) Fine-tuned ($D_i$) + Fine-tuned ($D_r$) : uses fine-tuned $f_i$ in round 1 and fine-tuned subsequent retriever $f_r$ in round 2. This is more costly as it requires two retrieval indexes and two retrieval models.





\section{Results}

\subsection{In-Domain Results}\label{subsec:in-domain-results}

\paragraph{Task Performance}
Table~\ref{tab:qampari-abs} reports retrieval performance on the QAMPARI test set. Fine-tuning with in-domain data improves raw performance significantly, especially improving weaker base retrievers (Contriever, Qwen) to be on-par with stronger base model (INF-Retriever). In-domain retrievers improve the agentic approaches, but overall, the agentic approaches underperform the baseline in all three base models, even when paired with the fine-tuned retriever. This could be caused by distribution shift from their query LLM training data, which typically requires multi-hop reasoning but not comprehensive answer coverage. 


\begin{table*}
\caption{\textbf{Out-of-Domain Generalization.} Results on QUEST (N=1727) and WebQuestionsSP (N=1639) test sets. We report MRecall@100 (MR) and Recall@100 (R). We report performances of different base retrieval models, agentic baselines, and RVR methods. For RVR, we use a verifier budget of 100. For our results, statistical significance is tested using bootstrap resampling with 10,000 trials at $\alpha = 0.05$. $\dagger$ indicates statistically significant improvement over Base. * indicates statistically significant improvement over Base + Base}
\label{tab:ood-generalization}
\vskip 0.15in
\begin{center}
\begin{small}
\setlength{\tabcolsep}{2pt}
\renewcommand{\arraystretch}{0.98}
\begin{tabular}{l cc|cc|cc|cc|cc|cc}
\toprule
& \multicolumn{6}{c}{\textbf{QUEST}} & \multicolumn{6}{c}{\textbf{WebQuestionsSP}} \\
\cmidrule(lr){2-7} \cmidrule(lr){8-13}
& \multicolumn{2}{c}{\textbf{Contriever}} &
\multicolumn{2}{c}{\textbf{Qwen3}} &
\multicolumn{2}{c}{\textbf{INF}} &
\multicolumn{2}{c}{\textbf{Contriever}} &
\multicolumn{2}{c}{\textbf{Qwen3}} &
\multicolumn{2}{c}{\textbf{INF}}  \\
\cmidrule(lr){2-3} \cmidrule(lr){4-5} \cmidrule(lr){6-7} \cmidrule(lr){8-9} \cmidrule(lr){10-11} \cmidrule(lr){12-13}
 & MR@100 & R@100 & MR@100 & R@100 & MR@100 & R@100 & MR@100 & R@100 & MR@100 & R@100 & MR@100 & R@100 \\ 
\midrule
Base &  3.24 & 23.79 & 3.13 & 21.38 & 4.75 & 26.60 & 62.16 & 77.38 & 61.00 & 76.06 & 62.47 & 77.39 \\
FT $(D_i)$ &  3.13 & 18.67 & 2.43 & 18.38 & 4.75 & 26.31 & 49.45 & 65.72 & 46.68 & 61.50 & 51.60 & 67.06 \\
\midrule
Tongyi & 0.93 & 10.51 & 2.49 & 19.52 & 3.30 & 21.95 & 52.33 & 65.84 & 57.56 & 71.80 & 58.29 & 72.37\\
SearchR1 & 0.81 & 9.54 & 3.19 & 20.83 & 3.53 & 23.01 & 54.91 & 68.92 & 61.49 & 76.67 & 61.98 & 76.82 \\
\midrule
\multicolumn{7}{l}{\textbf{Ours: Retrieve-Verify-Retrieve}}\\
Base + Base & 3.42 & 24.85$^{\dagger}$ & 3.30 & 22.67$^{\dagger}$ & 4.81 & 27.21$^{\dagger}$ & \textbf{62.96}$^{\dagger}$ & \textbf{77.93}$^{\dagger}$ & \textbf{62.53}$^{\dagger}$ & \textbf{77.54}$^{\dagger}$ & \textbf{63.21}$^{\dagger}$ & \textbf{78.20}$^{\dagger}$ \\
Base + FT ($D_r$) & \textbf{4.52}$^{\dagger*}$ & \textbf{26.01}$^{\dagger*}$ & \textbf{4.40}$^{\dagger*}$ & \textbf{25.43}$^{\dagger*}$ & \textbf{6.02}$^{\dagger*}$ & \textbf{30.53}$^{\dagger*}$ & 61.49 & 76.81 & 60.81 & 76.28 & 62.72 & 77.91\\
\bottomrule
\end{tabular}
\end{small}
\end{center}
\end{table*}

Our approach outperforms all baselines, raising both MRecall and R for all three base retrieval setting. Even using the same fine-tuned retriever $D_i$ as the baseline, verification step can bring gains when paired with stronger base retrievers (Qwen3, INF). Training retriever to collect complementary output $D_r$ further improves result. Having separate models (Fine-tuned $(D_i)$ + Fine-tuned $(D_r)$) further shows improvements. 
\paragraph{Efficiency}

We evaluate the efficiency on the QAMPARI test set across two dimensions: time and memory. For time, we report seconds per query (s/q) for both retrieval and verification/query generation components, measured with NVIDIA H200 GPUs. Retrieval time includes query encoding and k-nearest-neighbor search over the document index. Verification time depends on the verifier budget $B$ (number of documents verified per query) for our system, and the agentic search query generation process for agentic model. For memory, we report GPU memory usage in gigabytes for three components: the LLM verifier, the retrieval model, and the retrieval index. 
Table~\ref{tab:efficiency_time} reports the time efficiency across different retrieval models and configurations. The default single-pass outperforms others, while the agentic search is substantially slower. Two-round retrieval with verification incurs additional overhead that scales with the verifier budget and number of calls, making it 2-3 times slower than baseline. 

Table~\ref{tab:efficiency_memory} shows memory requirements, reporting both RVR and agentic search requires additional GPU memory compared to baseline. Having a separate model for initial retriever and subsequent retriever is more costly. 

\subsection{Generalization to Other Datasets} 

\paragraph{Setting}
We evaluate our models on two out-of-domain datasets: QUEST and WebQuestionsSP. We find that fine-tuned $f_i$ underperforms the baseline on these datasets, potentially due to domain shift. Therefore, we use the base retriever for the initial retrieval in our RVR framework. We evaluate two settings: (1) using the off-the-shelf retriever for both initial and subsequent retrieval (Base + Base), and (2) using the base retriever for initial retrieval and a subsequent retriever $D_r$ finetuned on QAMPARI (Base + FT($D_r$)). 

\paragraph{Results}
Table~\ref{tab:ood-generalization} reports results with $B$=100, while Table~\ref{tab:quest-abs} in the Appendix shows results with $B$=50. Our RVR framework largely outperforms the one-round baseline across both datasets. For QUEST, Base + FT($D_r$) achieves the strongest performance, demonstrating that a retriever trained to find complementary results generalizes well despite being fine-tuned on QAMPARI. For WebQuestionsSP, RVR still outperforms the baseline, though Base + FT($D_r$) performs slightly worse than Base + Base due to domain mismatch between QAMPARI and WebQuestionsSP.

\section{Analysis}

\subsection{The Impact of Verifier Performance}\label{subsec:verifier_performance}

\paragraph{Intrinsic Verifier Evaluation} 
The goal of verifier is to identify retrieved documents that are relevant to the original query. To evaluate their performance, we generate the following data: for each query in QAMPARI test set, we use three fine-tuned retrieval models $f_i$ from different base retrievers to retrieve 100 documents. We judge whether they contain gold answer using the gold label set. For this set, we observe 21.07\% being positive and 78.93\% being negative documents. 

We report the performance of three LLMs, GPT-5-nano, Qwen3-4B-Instruct-2507 \citep{qwen3technicalreport}, and Qwen3-30B-A3B-Instruct-2507 \citep{qwen3technicalreport}, as verifier on this dataset. Table~\ref{tab:verifier-baselines} presents the results. Qwen3-30B-A3B-Instruct-2507 model achieves the highest recall, which aligns with our objective of maximizing retrieval coverage. Therefore, we use this as the verifier for main experiments.

\begin{table}
\caption{Average verifier performance on top-100 retrieved documents on QAMPARI test set. Qwen3-30B-A3B-Instruct-2507 performs the best in terms of recall and is used as our verifier for the main experiments.}
\label{tab:verifier-baselines}
\begin{center}
\begin{small}
\begin{sc}
\setlength{\tabcolsep}{3pt}
\renewcommand{\arraystretch}{0.98}
\begin{tabular}{lccc}
\toprule
\textbf{Verifier} & \textbf{Precision} & \textbf{Recall} & \textbf{Accuracy} \\
\midrule
Random Baseline             & 6.71  & 6.42  & 64.85 \\
\midrule
GPT-5-nano                  & 50.25 & 54.70 & 82.29 \\
Qwen3-4B-Inst               & 43.62 & 52.57 & 77.63 \\
Qwen3-30B-Inst              & 34.06 & 74.05 & 68.43 \\
\bottomrule
\end{tabular}
\end{sc}
\end{small}
\end{center}
\vskip -0.1in
\end{table}

\begin{table}
\caption{Comparing the performance of using oracle vs. LLM verifier (Qwen3-30B) on QAMPARI test set. We evaluate MRecall@100 in the RVR setting FT $(D_i)$ + FT $(D_r)$, with a verifier budget of 100. Using LLM (Qwen3) verifier comes close to using an oracle verifier (upper bound) and outperforms \textsc{TopK} (baseline). }
\label{tab:oracle-verifier}
\begin{center}
\begin{small}
\begin{sc}
\setlength{\tabcolsep}{3pt}
\renewcommand{\arraystretch}{0.98}
\begin{tabular}{lccc}
\toprule
\textbf{Verifier} & \textbf{Contriever} & \textbf{Qwen3} & \textbf{INF} \\
\midrule
Oracle & \textbf{33.80}  &\textbf{33.60}  & \textbf{36.30}\\
LLM (Qwen3-30B) & 31.60 & 31.40 & 33.70 \\
TopK & 26.90 & 27.50  & 28.50 \\
\bottomrule
\end{tabular}
\end{sc}
\end{small}
\end{center}
\vskip -0.1in
\end{table}
\paragraph{Extrinsic Verifier Evaluation} In this section, we isolate the impact of verifier performance in end-to-end retrieval performance. We evaluate verifier performance by comparing against a baseline and an upper bound. As a baseline (TopK), we use the top $M$ documents ranked by the initial retriever to form the query context without a verifier. The final output set (K=100) combines the top 50 documents from the initial retriever with the top 50 non-duplicate documents from the subsequent retriever. \begin{figure*}
\begin{center}
\centerline{\includegraphics[width=\textwidth]{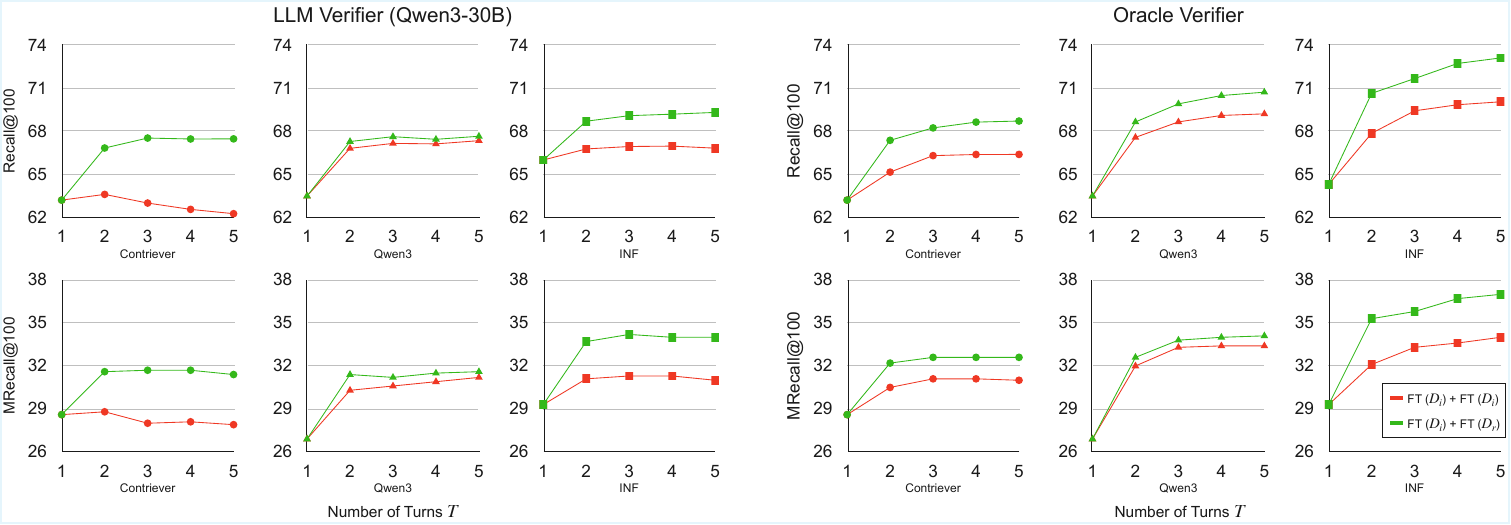}}
\caption{Multi-turn Generalization Results. This figure illustrates the change in Recall@100 and MRecall@100 across five iterations with a verifier budget of 100. Left panels show results with LLM verifier (Qwen3-30B), while right panels show results with oracle verifier that selects documents containing unique answer strings to be used as input. Performance with the LLM verifier plateaus after the second iteration, whereas the oracle verifier shows continued improvement, indicating substantial headroom for better verification mechanisms.}
\label{fig:varying-iterations}
\end{center}
\vskip -0.2in
\end{figure*}

As an upper bound, we use oracle verifier that has access to gold answer strings. For each example, we have a set of answer strings. Our oracle verifier leverages these answer strings to determine which documents are gold. If a document contains any of the answer strings, it is considered a gold document. 


Table~\ref{tab:oracle-verifier} presents the comparison across verifier settings: oracle, LLM (Qwen3-30B), and TopK. Across all base models, we see noticeable gains with oracle verifier, suggesting performance gain can be achieved by improving the verifier. 

\subsection{Multi-turn Generalization} 

Figure~\ref{fig:varying-iterations} compares performance across five iterations using both LLM and oracle verifiers. The left panel shows results with our LLM verifier (Qwen3-30B), where gains plateau after the second iteration. The right panel shows results with an oracle verifier that selects documents containing unique answer strings for $D_\text{ctx}$, demonstrating steady improvements across all five iterations. This disparity suggests the LLM verifier tends to select redundant documents in later iterations, even when the retrieved set contains documents with unique answers. These results indicate that improved verification mechanisms could substantially enhance RVR performance.


\subsection{The Impact of Verifier Budget}
\begin{figure}
\begin{center}
\centerline{\includegraphics[width=\columnwidth]{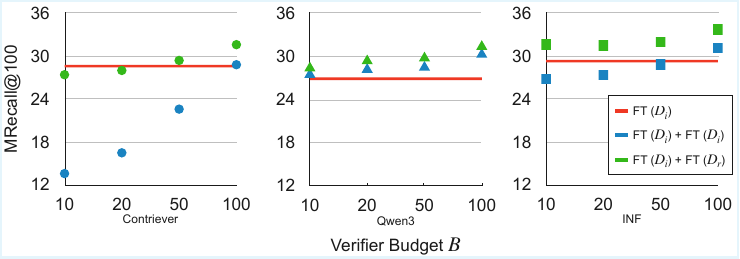}}
\caption{Varying the verifier budget. We evaluate MR@100 across three new verifier budgets on QAMPARI dataset. RVR is shown in green and blue, while our one-round baseline in red. See Appendix~\ref{appendix:varying-verifier-budget-precision-recall} for Recall@100 results.}
\label{fig:varying-verifier-budget}
\end{center}
\vskip -0.2in
\end{figure}

Figure~\ref{fig:varying-verifier-budget} plots the system performance for varying verifier budget (10, 20, 50, 100 documents) to evaluate models under more resource-constrained settings. The scores for 100 document setting is the same as reported in Table~\ref{tab:qampari-abs}.

We compare two settings of our Retrieve-Verify-Retrieve, one with FT($D_i$) and another with FT($D_i$) and FT($D_r$). Across all models, the absolute performance decreases as the verifier budget shrinks. Using fine-tuned subsequent retriever FT($D_r$) was particularly helpful in more restrictive budget setting. 

\subsection{The Impact of Varying Input Length}

\begin{figure}
\begin{center}
\centerline{\includegraphics[width=0.7\columnwidth]{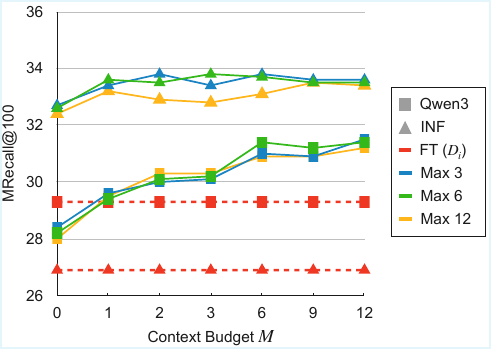}}
\caption{The performance (MRecall@100) with varying number of input documents at inference time (Context Budget $M$). We compare models fine-tuned with different maximum document counts (3, 6, and 12 docs) for INF and Qwen3. Different colors denote fine-tuning with different number of documents, and shapes indicate the retrievers used. Other metric results (Recall@100) are provided in Appendix~\ref{appendix:varying-input}. }
\label{fig:varying-input-precision-mrecall}
\end{center}
\vskip -0.2in
\end{figure}

We study the impact of varying the maximum number of input documents provided to the subsequent retriever. Here, we do not report results for Contriever, as it has a sequence length limit of 512 tokens, resulting in truncation when we provide more than $3$ input documents. We run inference with our RVR model (Fine-tuned + Ours) setting, fine-tuned with a maximum of 3, 6, and 12 input documents. We further vary the context budget, $M$, used during inference time. 

Figure~\ref{fig:vary-input-recall} displays the trends, as we sample from 0 up to 12 input documents. We find that increasing the number of documents at inference time beyond 6 does not show strong gains. We also observe that on average, models fine-tuned with up to $6$ input documents perform better. 
\subsection{Ablation Studies on Retrievers}
Table~\ref{tab:ours-ablation} presents ablation study of retriever components. In rows 1 and 2, we report using off-the-shelf base retriever instead of fine-tuned retriever for $f_i$. As long as we fine-tune the subsequent retriever, the performance is quite competitive, sometimes even outperforming setting with both initial and subsequent retriever fine-tuned.

\begin{table}
\caption{Ablation study on QAMPARI test set. Each row represents a RVR configuration, denoted in the format of $f_i$ + $f_r$ (initial + subsequent retriever). \textbf{Base}: off-the-shelf pretrained retriever; \textbf{FT($D_i$)}: fine-tuned on initial retrieval data; \textbf{FT($D_r$)}: fine-tuned on subsequent retrieval data with document context; \textbf{FT($D_i$ + $D_r$)}: fine-tuned on both objectives jointly.}
\vspace{-0.5em}
\label{tab:ours-ablation}
\begin{center}
\begin{small}
\setlength{\tabcolsep}{2.8pt}
\renewcommand{\arraystretch}{1.0}
\begin{tabular}{l cc|cc|cc}
\toprule
&\multicolumn{2}{c|}{\textbf{Contriever}} & \multicolumn{2}{c|}{\textbf{Qwen3}}  & \multicolumn{2}{c}{\textbf{INF}}  \\ 
\cmidrule(lr){2-3} \cmidrule(lr){4-5} \cmidrule(lr){6-7}
& MR & R& MR & R& MR & R\\ \midrule
Base + Base&21.30 & 58.10 &  20.20 & 58.12 & 28.50 & 65.29\\
FT($D_i$) + FT($D_i$)&  28.80 & 63.59 & 30.30 & 66.80  & 31.10 & 66.76 \\ 
FT($D_i$ + $D_r$) & \multirow{2}{*}{\textbf{31.70}} & \multirow{2}{*}{66.12} & \multirow{2}{*}{26.30} & \multirow{2}{*}{61.24} & \multirow{2}{*}{32.40} & \multirow{2}{*}{68.04} \\
\hspace{0.2em} + FT($D_i$ + $D_r$) & & & & & & \\
\midrule
Base + FT ($D_r$) & 30.70 & \textbf{67.25}  & 29.20 & 66.36 & \textbf{34.20} & \textbf{69.29}\\
FT ($D_i$) + FT ($D_r$) &{31.60} & 66.83  & \textbf{31.40} & \textbf{67.28} &  33.70 & 68.70 \\
\bottomrule
\end{tabular}
\end{small}
\end{center}
\vskip -0.1in
\end{table}

\subsection{The contribution of 1st turn vs. 2nd turn retrieval}
We analyze the individual contributions of our first-stage (base) and second-stage (iterative) retrievers. Table~\ref{tab:stage-contribution-v100} presents a breakdown across our three models with $B$=100. Table~\ref{tab:stage-contribution} shows results for $B$=50.

\begin{table}
\caption{Contribution analysis of first and second stage retrieval on QAMPARI test set. We report the average number of new gold documents retrieved and the average number of new answers covered per question in each stage, with verifier budget $B$=100.}
\vspace{-1em}
\label{tab:stage-contribution-v100}
\vskip 0.15in
\begin{center}
\begin{small}
\setlength{\tabcolsep}{3pt}
\renewcommand{\arraystretch}{1.0}
\begin{tabular}{l cc|cc}
\toprule
&  \multicolumn{2}{c|}{\textbf{\# New Gold Docs.}} & \multicolumn{2}{c}{\textbf{\# New Uniq. Ans.}} \\
\cmidrule(lr){2-3} \cmidrule(lr){4-5}
& Turn 1 & Turn 2 & Turn 1 & Turn 2 \\
\midrule
\multicolumn{4}{l}{Contriever} \\
 FT ($D_i$) + FT ($D_i$) & 49.59 & 17.60 & 7.12 & 0.33 \\
                          FT ($D_i$) + FT ($D_r$) & 49.59 & \textbf{22.83} & 7.12 & \textbf{0.73} \\
\midrule
\multicolumn{4}{l}{Qwen3}\\
FT ($D_i$) + FT ($D_i$) & 50.94 & 22.35 & 7.17 & 0.66 \\
                        FT ($D_i$) + FT ($D_r$) & 50.94 & \textbf{23.07} & 7.17 & \textbf{0.75} \\
\midrule
\multicolumn{4}{l}{INF}\\
 FT ($D_i$) + FT ($D_i$) & 55.32 & 22.73 & 7.50 & 0.48 \\
                        FT ($D_i$) + FT ($D_r$) & 55.32 & \textbf{26.12} & 7.50 & \textbf{0.72} \\
\bottomrule
\end{tabular}
\end{small}
\end{center}
\vskip -0.1in
\end{table}

The first-stage retriever provides the initial set of relevant documents. Across all models, the first stage contributes approximately 50 gold documents, corresponding to 7 unique answers, per question on average, demonstrating that initial retriever already captures a substantial portion of relevant information. The second-stage iterative retriever, conditioned on verified documents from the first stage, contributes an additional 22-26 gold documents, up to 1 unique answer, per question. This consistent improvement demonstrates that conditioning on previously retrieved context enables the model to discover initially missed documents.

\section{Conclusion}
We introduced Retrieve-Verify-Retrieve framework, a framework that conditions on previously retrieved evidence and explicitly optimizes for answer coverage. By training retrievers to predict missing gold documents and integrating verifier-guided iteration, our approach consistently expands retrieval coverage while reducing redundancy. Results on QAMPARI and zero-shot generalization to QUEST demonstrate that iterative conditioning provides a robust and general mechanism for improving retrieval completeness.




\section*{Impact Statement}
This paper introduces a retrieval framework designed to improve answer coverage and reduce redundancy in information retrieval systems. By enabling retrieval models to iteratively reason over previously retrieved evidence, the proposed approach may improve the completeness and reliability of systems used in applications such as search engines, question answering, and knowledge discovery. 

Potential positive impacts include more thorough information access, reduced bias toward dominant viewpoints, and improved support for users seeking comprehensive answers. As with any retrieval technology, the quality of results ultimately depend on the underlying data sources and the design of downstream systems. Potential risks mirror those of existing retrieval systems, including surfacing misinformation or amplifying biases present in the corpus.



\section*{Acknowledgements}
We thank NYU NLP group for valuable feedback, especially Fangyuan Xu. This work was supported in part through
the NYU IT High Performance Computing resources, services, and staff expertise. The work is partially funded by NSF CAREER award 2443271 and NSF award RI-2521091.

\bibliography{example_paper}

@misc{chen2025browsecompplusfairtransparentevaluation,
      title={BrowseComp-Plus: A More Fair and Transparent Evaluation Benchmark of Deep-Research Agent}, 
      author={Zijian Chen and Xueguang Ma and Shengyao Zhuang and Ping Nie and Kai Zou and Andrew Liu and Joshua Green and Kshama Patel and Ruoxi Meng and Mingyi Su and Sahel Sharifymoghaddam and Yanxi Li and Haoran Hong and Xinyu Shi and Xuye Liu and Nandan Thakur and Crystina Zhang and Luyu Gao and Wenhu Chen and Jimmy Lin},
      year={2025},
      eprint={2508.06600},
      archivePrefix={arXiv},
      primaryClass={cs.CL},
      url={https://arxiv.org/abs/2508.06600}, 
}

@inproceedings{fang2025kiragknowledgedriveniterativeretriever,
    title = "{K}i{RAG}: Knowledge-Driven Iterative Retriever for Enhancing Retrieval-Augmented Generation",
    author = "Fang, Jinyuan  and
      Meng, Zaiqiao  and
      MacDonald, Craig",
    editor = "Che, Wanxiang  and
      Nabende, Joyce  and
      Shutova, Ekaterina  and
      Pilehvar, Mohammad Taher",
    booktitle = "Proceedings of the 63rd Annual Meeting of the Association for Computational Linguistics (Volume 1: Long Papers)",
    month = jul,
    year = "2025",
    address = "Vienna, Austria",
    publisher = "Association for Computational Linguistics",
    url = "https://aclanthology.org/2025.acl-long.929/",
    doi = "10.18653/v1/2025.acl-long.929",
    pages = "18969--18985",
}

@inproceedings{dpr,
    title = "Dense Passage Retrieval for Open-Domain Question Answering",
    author = "Karpukhin, Vladimir  and
      Oguz, Barlas  and
      Min, Sewon  and
      Lewis, Patrick  and
      Wu, Ledell  and
      Edunov, Sergey  and
      Chen, Danqi  and
      Yih, Wen-tau",
    editor = "Webber, Bonnie  and
      Cohn, Trevor  and
      He, Yulan  and
      Liu, Yang",
    booktitle = "Proceedings of the 2020 Conference on Empirical Methods in Natural Language Processing (EMNLP)",
    month = nov,
    year = "2020",
    address = "Online",
    publisher = "Association for Computational Linguistics",
    url = "https://aclanthology.org/2020.emnlp-main.550/",
    doi = "10.18653/v1/2020.emnlp-main.550",
    pages = "6769--6781"
}

@inproceedings{
asai2024selfrag,
title={Self-{RAG}: Learning to Retrieve, Generate, and Critique through Self-Reflection},
author={Akari Asai and Zeqiu Wu and Yizhong Wang and Avirup Sil and Hannaneh Hajishirzi},
booktitle={The Twelfth International Conference on Learning Representations},
year={2024},
url={https://openreview.net/forum?id=hSyW5go0v8}
}

@article{
izacard2021unsupervised,
title={Unsupervised Dense Information Retrieval with Contrastive Learning},
author={Gautier Izacard and Mathilde Caron and Lucas Hosseini and Sebastian Riedel and Piotr Bojanowski and Armand Joulin and Edouard Grave},
journal={Transactions on Machine Learning Research},
issn={2835-8856},
year={2022},
url={https://openreview.net/forum?id=jKN1pXi7b0},
note={}
}

@inproceedings{qi2019iterativequery,
    title = "Answering Complex Open-domain Questions Through Iterative Query Generation",
    author = "Qi, Peng  and
      Lin, Xiaowen  and
      Mehr, Leo  and
      Wang, Zijian  and
      Manning, Christopher D.",
    editor = "Inui, Kentaro  and
      Jiang, Jing  and
      Ng, Vincent  and
      Wan, Xiaojun",
    booktitle = "Proceedings of the 2019 Conference on Empirical Methods in Natural Language Processing and the 9th International Joint Conference on Natural Language Processing (EMNLP-IJCNLP)",
    month = nov,
    year = "2019",
    address = "Hong Kong, China",
    publisher = "Association for Computational Linguistics",
    url = "https://aclanthology.org/D19-1261/",
    doi = "10.18653/v1/D19-1261",
    pages = "2590--2602",
}

@inproceedings{trivedi2023ircot,
    title = "Interleaving Retrieval with Chain-of-Thought Reasoning for Knowledge-Intensive Multi-Step Questions",
    author = "Trivedi, Harsh  and
      Balasubramanian, Niranjan  and
      Khot, Tushar  and
      Sabharwal, Ashish",
    editor = "Rogers, Anna  and
      Boyd-Graber, Jordan  and
      Okazaki, Naoaki",
    booktitle = "Proceedings of the 61st Annual Meeting of the Association for Computational Linguistics (Volume 1: Long Papers)",
    month = jul,
    year = "2023",
    address = "Toronto, Canada",
    publisher = "Association for Computational Linguistics",
    url = "https://aclanthology.org/2023.acl-long.557/",
    doi = "10.18653/v1/2023.acl-long.557",
    pages = "10014--10037",
}

@inproceedings{min2021jointpr,
    title = "Joint Passage Ranking for Diverse Multi-Answer Retrieval",
    author = "Min, Sewon  and
      Lee, Kenton  and
      Chang, Ming-Wei  and
      Toutanova, Kristina  and
      Hajishirzi, Hannaneh",
    editor = "Moens, Marie-Francine  and
      Huang, Xuanjing  and
      Specia, Lucia  and
      Yih, Scott Wen-tau",
    booktitle = "Proceedings of the 2021 Conference on Empirical Methods in Natural Language Processing",
    month = nov,
    year = "2021",
    address = "Online and Punta Cana, Dominican Republic",
    publisher = "Association for Computational Linguistics",
    url = "https://aclanthology.org/2021.emnlp-main.560/",
    doi = "10.18653/v1/2021.emnlp-main.560",
    pages = "6997--7008",
}

@inproceedings{dhuliawala2024chainofverification,
    title = "Chain-of-Verification Reduces Hallucination in Large Language Models",
    author = "Dhuliawala, Shehzaad  and
      Komeili, Mojtaba  and
      Xu, Jing  and
      Raileanu, Roberta  and
      Li, Xian  and
      Celikyilmaz, Asli  and
      Weston, Jason",
    editor = "Ku, Lun-Wei  and
      Martins, Andre  and
      Srikumar, Vivek",
    booktitle = "Findings of the Association for Computational Linguistics: ACL 2024",
    month = aug,
    year = "2024",
    address = "Bangkok, Thailand",
    publisher = "Association for Computational Linguistics",
    url = "https://aclanthology.org/2024.findings-acl.212/",
    doi = "10.18653/v1/2024.findings-acl.212",
    pages = "3563--3578"
}

@inproceedings{yang2018hotpotqa,
  title={HotpotQA: A dataset for diverse, explainable multi-hop question answering},
  author={Yang, Zhilin and Qi, Peng and Zhang, Saizheng and Bengio, Yoshua and Cohen, William and Salakhutdinov, Ruslan and Manning, Christopher D},
  booktitle={Proceedings of the 2018 conference on empirical methods in natural language processing},
  pages={2369--2380},
  year={2018}
}

@inproceedings{malaviya2023quest,
    title = "{QUEST}: A Retrieval Dataset of Entity-Seeking Queries with Implicit Set Operations",
    author = "Malaviya, Chaitanya  and
      Shaw, Peter  and
      Chang, Ming-Wei  and
      Lee, Kenton  and
      Toutanova, Kristina",
    editor = "Rogers, Anna  and
      Boyd-Graber, Jordan  and
      Okazaki, Naoaki",
    booktitle = "Proceedings of the 61st Annual Meeting of the Association for Computational Linguistics (Volume 1: Long Papers)",
    month = jul,
    year = "2023",
    address = "Toronto, Canada",
    publisher = "Association for Computational Linguistics",
    url = "https://aclanthology.org/2023.acl-long.784/",
    doi = "10.18653/v1/2023.acl-long.784",
    pages = "14032--14047",
}

@article{chen2025beyond,
  title={Beyond Single Embeddings: Capturing Diverse Targets with Multi-Query Retrieval},
  author={Chen, Hung-Ting and Liu, Xiang and Ravfogel, Shauli and Choi, Eunsol},
  journal={arXiv preprint arXiv:2511.02770},
  year={2025}
}

@article{qwen3embedding,
  title={Qwen3 Embedding: Advancing Text Embedding and Reranking Through Foundation Models},
  author={Zhang, Yanzhao and Li, Mingxin and Long, Dingkun and Zhang, Xin and Lin, Huan and Yang, Baosong and Xie, Pengjun and Yang, An and Liu, Dayiheng and Lin, Junyang and Huang, Fei and Zhou, Jingren},
  journal={arXiv preprint arXiv:2506.05176},
  year={2025}
}

@misc {infly-ai_2025,
    author       = {Yang, Junhan and Wan, Jiahe and Yao, Yichen and Chu, Wei and Xu, Yinghui and Wang, Emma and Qi, Yuan},
    title        = { inf-retriever-v1 (Revision 5f469d7) },
    year         = 2025,
    url          = { https://huggingface.co/infly/inf-retriever-v1 },
    doi          = { 10.57967/hf/4262 },
    publisher    = { Hugging Face }
}

@misc{qwen3technicalreport,
      title={Qwen3 Technical Report}, 
      author={Qwen Team},
      year={2025},
      eprint={2505.09388},
      archivePrefix={arXiv},
      primaryClass={cs.CL},
      url={https://arxiv.org/abs/2505.09388}, 
}

@inproceedings{katz-etal-2023-neretrieve,
    title = "{NER}etrieve: Dataset for Next Generation Named Entity Recognition and Retrieval",
    author = "Katz, Uri  and
      Vetzler, Matan  and
      Cohen, Amir  and
      Goldberg, Yoav",
    editor = "Bouamor, Houda  and
      Pino, Juan  and
      Bali, Kalika",
    booktitle = "Findings of the Association for Computational Linguistics: EMNLP 2023",
    month = dec,
    year = "2023",
    address = "Singapore",
    publisher = "Association for Computational Linguistics",
    url = "https://aclanthology.org/2023.findings-emnlp.218/",
    doi = "10.18653/v1/2023.findings-emnlp.218",
    pages = "3340--3354",
    
}

@inproceedings{
loshchilov2018decoupled,
title={Decoupled Weight Decay Regularization},
author={Ilya Loshchilov and Frank Hutter},
booktitle={International Conference on Learning Representations},
year={2019},
url={https://openreview.net/forum?id=Bkg6RiCqY7},
}

@article{schulman2017proximal,
  title={Proximal policy optimization algorithms},
  author={Schulman, John and Wolski, Filip and Dhariwal, Prafulla and Radford, Alec and Klimov, Oleg},
  journal={arXiv preprint arXiv:1707.06347},
  year={2017}
}

@article{kwiatkowski-etal-2019-natural,
    title = "Natural Questions: A Benchmark for Question Answering Research",
    author = "Kwiatkowski, Tom  and
      Palomaki, Jennimaria  and
      Redfield, Olivia  and
      Collins, Michael  and
      Parikh, Ankur  and
      Alberti, Chris  and
      Epstein, Danielle  and
      Polosukhin, Illia  and
      Devlin, Jacob  and
      Lee, Kenton  and
      Toutanova, Kristina  and
      Jones, Llion  and
      Kelcey, Matthew  and
      Chang, Ming-Wei  and
      Dai, Andrew M.  and
      Uszkoreit, Jakob  and
      Le, Quoc  and
      Petrov, Slav",
    editor = "Lee, Lillian  and
      Johnson, Mark  and
      Roark, Brian  and
      Nenkova, Ani",
    journal = "Transactions of the Association for Computational Linguistics",
    volume = "7",
    year = "2019",
    address = "Cambridge, MA",
    publisher = "MIT Press",
    url = "https://aclanthology.org/Q19-1026/",
    doi = "10.1162/tacl_a_00276",
    pages = "452--466",
    
}

@article{Asai2022TaskawareRW,
  title={Task-aware Retrieval with Instructions},
  author={Akari Asai and Timo Schick and Patrick Lewis and Xilun Chen and Gautier Izacard and Sebastian Riedel and Hannaneh Hajishirzi and Wen-tau Yih},
  journal={ArXiv},
  year={2022},
  volume={abs/2211.09260},
  url={https://api.semanticscholar.org/CorpusID:253581733}
}

@inproceedings{
xiong2020multihop,
title={Answering Complex Open-Domain Questions with Multi-Hop Dense Retrieval},
author={Wenhan Xiong and Xiang Li and Srini Iyer and Jingfei Du and Patrick Lewis and William Yang Wang and Yashar Mehdad and Scott Yih and Sebastian Riedel and Douwe Kiela and Barlas Oguz},
booktitle={International Conference on Learning Representations},
year={2021},
url={https://openreview.net/forum?id=EMHoBG0avc1}
}

@inproceedings{amouyal2022qampari,
    title = "{QAMPARI}: A Benchmark for Open-domain Questions with Many Answers",
    author = "Amouyal, Samuel  and
      Wolfson, Tomer  and
      Rubin, Ohad  and
      Yoran, Ori  and
      Herzig, Jonathan  and
      Berant, Jonathan",
    editor = "Gehrmann, Sebastian  and
      Wang, Alex  and
      Sedoc, Jo{\~a}o  and
      Clark, Elizabeth  and
      Dhole, Kaustubh  and
      Chandu, Khyathi Raghavi  and
      Santus, Enrico  and
      Sedghamiz, Hooman",
    booktitle = "Proceedings of the Third Workshop on Natural Language Generation, Evaluation, and Metrics (GEM)",
    month = dec,
    year = "2023",
    address = "Singapore",
    publisher = "Association for Computational Linguistics",
    url = "https://aclanthology.org/2023.gem-1.9/",
    pages = "97--110",
}

@inproceedings{chen-choi-2025-open,
    title = "Open-World Evaluation for Retrieving Diverse Perspectives",
    author = "Chen, Hung-Ting  and
      Choi, Eunsol",
    editor = "Chiruzzo, Luis  and
      Ritter, Alan  and
      Wang, Lu",
    booktitle = "Proceedings of the 2025 Conference of the Nations of the Americas Chapter of the Association for Computational Linguistics: Human Language Technologies (Volume 1: Long Papers)",
    month = apr,
    year = "2025",
    address = "Albuquerque, New Mexico",
    publisher = "Association for Computational Linguistics",
    url = "https://aclanthology.org/2025.naacl-long.431/",
    doi = "10.18653/v1/2025.naacl-long.431",
    pages = "8508--8528",
    ISBN = "979-8-89176-189-6",
}

@inproceedings{jin2025search,
  title={Search-r1: Training llms to reason and leverage search engines with reinforcement learning},
  author={Jin, Bowen and Zeng, Hansi and Yue, Zhenrui and Yoon, Jinsung and Arik, Sercan and Wang, Dong and Zamani, Hamed and Han, Jiawei},
  booktitle="Conference on Language Modeling",
  year={2025}
}

@article{tongyidr,
  title={Tongyi DeepResearch Technical Report},
  author={Team, Tongyi DeepResearch and Li, Baixuan and Zhang, Bo and Zhang, Dingchu and Huang, Fei and Li, Guangyu and Chen, Guoxin and Yin, Huifeng and Wu, Jialong and Zhou, Jingren and others},
  journal={arXiv preprint arXiv:2510.24701},
  year={2025}
}

@article{trivedi2022musique,
  title={MuSiQue: Multi-hop Questions via Single-hop Question Composition},
  author={Trivedi, Harsh and Balasubramanian, Niranjan and Khot, Tushar and Sabharwal, Ashish},
  journal={Transactions of the Association for Computational Linguistics},
  volume={10},
  pages={539--554},
  year={2022}
}

@article{shao2025dr,
  title={Dr tulu: Reinforcement learning with evolving rubrics for deep research},
  author={Shao, Rulin and Asai, Akari and Shen, Shannon Zejiang and Ivison, Hamish and Kishore, Varsha and Zhuo, Jingming and Zhao, Xinran and Park, Molly and Finlayson, Samuel G and Sontag, David and others},
  journal={arXiv preprint arXiv:2511.19399},
  year={2025}
}

@inproceedings{kwon2023efficient,
  title={Efficient Memory Management for Large Language Model Serving with PagedAttention},
  author={Woosuk Kwon and Zhuohan Li and Siyuan Zhuang and Ying Sheng and Lianmin Zheng and Cody Hao Yu and Joseph E. Gonzalez and Hao Zhang and Ion Stoica},
  booktitle={Proceedings of the ACM SIGOPS 29th Symposium on Operating Systems Principles},
  year={2023}
}

@article{liu2026agentic,
  title={Agentic-R: Learning to Retrieve for Agentic Search},
  author={Liu, Wenhan and Ma, Xinyu and Zhu, Yutao and Li, Yuchen and Shi, Daiting and Yin, Dawei and Dou, Zhicheng},
  journal={arXiv preprint arXiv:2601.11888},
  year={2026}
}

@inproceedings{yih-etal-2016-value,
    title = "The Value of Semantic Parse Labeling for Knowledge Base Question Answering",
    author = "Yih, Wen-tau  and
      Richardson, Matthew  and
      Meek, Chris  and
      Chang, Ming-Wei  and
      Suh, Jina",
    editor = "Erk, Katrin  and
      Smith, Noah A.",
    booktitle = "Proceedings of the 54th Annual Meeting of the Association for Computational Linguistics (Volume 2: Short Papers)",
    month = aug,
    year = "2016",
    address = "Berlin, Germany",
    publisher = "Association for Computational Linguistics",
    url = "https://aclanthology.org/P16-2033/",
    doi = "10.18653/v1/P16-2033",
    pages = "201--206"
}

@inproceedings{zhu-etal-2024-fanoutqa,
    title = "{F}an{O}ut{QA}: A Multi-Hop, Multi-Document Question Answering Benchmark for Large Language Models",
    author = "Zhu, Andrew  and
      Hwang, Alyssa  and
      Dugan, Liam  and
      Callison-Burch, Chris",
    editor = "Ku, Lun-Wei  and
      Martins, Andre  and
      Srikumar, Vivek",
    booktitle = "Proceedings of the 62nd Annual Meeting of the Association for Computational Linguistics (Volume 2: Short Papers)",
    month = aug,
    year = "2024",
    address = "Bangkok, Thailand",
    publisher = "Association for Computational Linguistics",
    url = "https://aclanthology.org/2024.acl-short.2/",
    doi = "10.18653/v1/2024.acl-short.2",
    pages = "18--37",
}

@article{sharifymoghaddam2026rerank,
  title={Rerank Before You Reason: Analyzing Reranking Tradeoffs through Effective Token Cost in Deep Search Agents},
  author={Sharifymoghaddam, Sahel and Lin, Jimmy},
  journal={arXiv preprint arXiv:2601.14224},
  year={2026}
}
\bibliographystyle{icml2026}

\newpage
\appendix
\onecolumn


\section{Datasets}

\subsection{QAMPARI}
\label{appendix:qampari-stat}
Table~\ref{tab:qampari-dataset-stats} displays our dataset statistics in detail. An example of a gold document from our QAMPARI test set, along with the query and answer strings is below.

\textbf{Query}: What manga was drawn by Ryoichi Ikegami?

\textbf{Answer Strings}: Heat, \textbf{\textit{Mai, the Psychic Girl}}, Wounded Man, Sanctuary, Crying Freeman, Strain

\textbf{Document Text}: \textbf{\textit{Mai, the Psychic Girl}}, known simply as in Japan, is a manga written by Kazuya Kudo and illustrated by Ryoichi Ikegami. The main character is Mai Kuju, a 14-year-old Japanese girl with powerful psychic abilities. She is being pursued by the Wisdom Alliance, an organization which secretly strives to control the world. The alliance already controls four other powerful psychic children, and it has hired the Kaieda Intelligence Agency to capture Mai. Media. Manga. \textbf{\textit{Mai, the Psychic Girl}} is one of the first manga series to be fully published in English.

\subsection{QUEST}
\label{appendix:quest-stat}
Below, we show an example query from the QUEST test set, its associated answer string, and a document from our corpus that contains this answer. Table~\ref{tab:qampari-dataset-stats} shows additional dataset statistics. 

\textbf{Query}: what are some 1950s comedy mystery or spy comedy films

\textbf{Answer Strings}: The Hole (1957 film), Charade (1953 film), Hot Stuff (1956 film), The Trouble with Harry, Father Brown (film), The Runaway Bus, Boston Quackie, \textbf{\textit{My Favorite Spy}}, The Fuller Brush Girl, Scared Stiff (1953 film), Our Man in Havana (film), Spy Chasers, Mrs. O'Malley and Mr. Malone, Clipped Wings (1953 film), Commotion on the Ocean, Down Among the Z Men, Knock on Wood (film), Top Secret (1952 film)

\textbf{Document Text}: 
\textbf{\textit{My Favorite Spy}} is a 1951 comedy film directed by Norman Z. McLeod and starring Bob Hope and Hedy Lamarr. Plot. US intelligence agents recruit burlesque comic Peanuts White to pose as international spy Eric Augustine, whom he resembles, to acquire a million-dollar microfilm in Tangier. There, he encounters the irresistible Lily Dalbray, Augustine's one-time friend, who is now in league with his arch-enemy, Brubaker.

\subsection{WebQuestionsSP}
\label{appendix:webqsp-stat}
Below, we show an example query from the WebQuestionsSP test set, its associated answer string, and a document from our corpus that contains this answer. Table~\ref{tab:qampari-dataset-stats} shows additional dataset statistics. 

\textbf{Query}: what does jamaican people speak

\textbf{Answer Strings}: \textbf{\textit{Jamaican English}}, Jamaican Creole English Language

\textbf{Document Text}: In \textbf{\textit{Jamaican English}}, normally reduced English vowels are sometimes not reduced, and other times are hyper-reduced, so that \"token\" is not but, yet \"cement\" can be as reduced as; the exact nuances of the rules at play here are also highly debated. Language use: Jamaican Standard English versus Patois. Jamaican Standard English and Jamaican Patois exist together in a post-creole speech continuum. Jamaican (Creole/Patois) is used by most people for everyday, informal situations - it is the language most Jamaicans use at home and are most familiar with, as well as the language of most local popular music.

\begin{table}
\caption{Dataset statistics showing the number of questions split across train, development, and test sets, along with the average number of answers and gold documents per question.}
\label{tab:qampari-dataset-stats}
\vskip 0.15in
\begin{center}
\begin{small}
\setlength{\tabcolsep}{4pt}
\renewcommand{\arraystretch}{0.98}
\begin{tabular}{lccccc}
\toprule
& \multicolumn{3}{c}{\textbf{\# examples}} & \textbf{\# answers} & \textbf{\# gold docs} \\
\cmidrule(lr){2-4} \cmidrule(lr){5-5} \cmidrule(lr){6-6}
\textbf{Dataset} & Train & Dev & Test & Avg. & Avg. \\ 
\midrule
QAMPARI & 34,425 & 6,191 & 1,000 & 14.43 & 16.58 \\
QUEST & --- & --- & 1,727 &  10.50 & --- \\
WebQuestionsSP & --- & --- & 1,639 & 8.75 & ---\\
\bottomrule
\end{tabular}
\end{small}
\end{center}
\end{table}


\subsection{Instruction Used for Instruction Tuned Models}
\label{appendix:instruction}
We use the same instruction to finetune our Qwen3-0.6B-Embed and inf-retriever-1.5B models: \textit{Given a query, retrieve relevant passages that answer the query}. We use this instruction during both fine-tuning and inference.

\section{More Analysis}

\subsection{Impact of Verifier Budget on Recall@K}
\label{appendix:varying-verifier-budget-precision-recall}
\begin{figure}
\begin{center}
\centerline{\includegraphics[width=0.7\columnwidth]{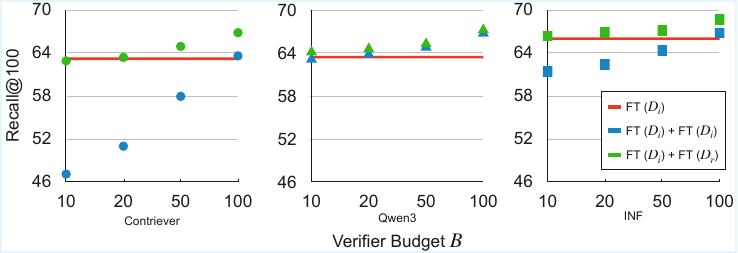}}
\caption{This figure shows the results of lowering verifier budget beyond 100 for Recall@100.}
\label{fig:varying-verifier-budget-precision-recall}
\end{center}
\vskip -0.2in
\end{figure}

Figure~\ref{fig:varying-verifier-budget-precision-recall} displays Precision@100 and Recall@100 with verifier budgets of 10, 20, 50, and 100.

\subsection{Impact of Varying Input Documents on MRecall@K}
\label{appendix:varying-input}
Figure~\ref{fig:varying-input-precision-mrecall} displays additional results for our experiment where we vary the number of input documents at inference time. 



\begin{figure}
\begin{center}
\centerline{\includegraphics[width=0.4\columnwidth]{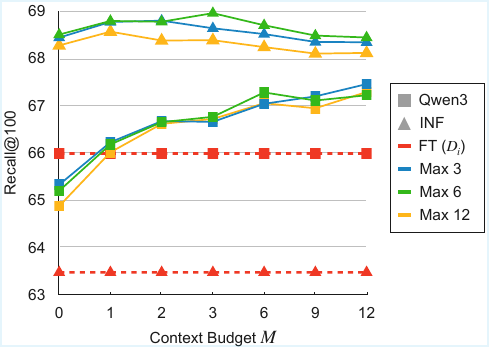}}
\caption{This figure shows the results of increasing the number of input documents on Recall@100 for INF and Qwen3 models fine-tuned with different maximum document counts: 3, 6, and 12. We evaluate across 7 different context budgets.}
\label{fig:vary-input-recall}
\end{center}
\vskip -0.2in
\end{figure}

\begin{figure}
\begin{center}
\centerline{\includegraphics[width=0.7\columnwidth]{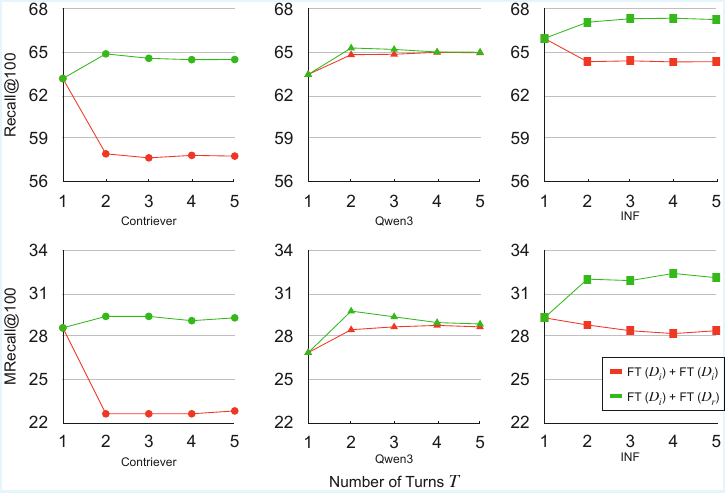}}
\caption{Multi-turn Generalization Results. This figure illustrates the change in Recall@100 and MRecall@100 across five iterations with a verifier budget of 50.}
\label{fig:varying-iterations-b50}
\end{center}
\vskip -0.2in
\end{figure}

\begin{table}
\caption{Contribution analysis of first-stage and second-stage retrieval on QAMPARI test set. We report the average number of gold documents retrieved and the average number of new answers covered per question in each stage with verifier budget of 50.}
\label{tab:stage-contribution}
\vskip 0.15in
\begin{center}
\begin{small}
\setlength{\tabcolsep}{3pt}
\renewcommand{\arraystretch}{1.0}
\begin{tabular}{ll cc|cc}
\toprule
& & \multicolumn{2}{c|}{\textbf{\# New Gold Docs.}} & \multicolumn{2}{c}{\textbf{\# New Uniq. Ans.}} \\
\cmidrule(lr){3-4} \cmidrule(lr){5-6}
& & Turn 1 & Turn 2 & Turn 1 & Turn 2 \\
\midrule
{Contr-} & FT ($D_i$) + FT ($D_i$) & 27.93 & 10.69 & 5.68 & 0.51 \\
                    iever        & FT ($D_i$) + FT ($D_r$) & 27.93 & \textbf{13.98} & 5.68 & \textbf{1.05} \\
\midrule
\multirow{2}{*}{Qwen3} & FT ($D_i$) + FT ($D_i$) & 28.95 & 13.53 & 5.84 & 0.89 \\
                       & FT ($D_i$) + FT ($D_r$) & 28.95 & \textbf{13.61} & 5.84 & \textbf{0.93} \\
\midrule
\multirow{2}{*}{INF} & FT ($D_i$) + FT ($D_i$)& 30.98 & 13.50 & 6.13 & 0.66 \\
                       & FT ($D_i$) + FT ($D_r$) & 30.98 & \textbf{15.09} & 6.13 & \textbf{1.04} \\
\bottomrule
\end{tabular}
\end{small}
\end{center}
\vskip -0.1in
\end{table}

\begin{table}
\caption{Comparing the performance of using oracle verifier vs. LLM verifier on QAMPARI test set. We evaluate MRecall@100 in RVR setting FT $(D_i)$ + FT $(D_r)$ with a verifier budget of 50. }
\label{tab:oracle-verifier-v50}
\begin{center}
\begin{small}
\begin{sc}
\setlength{\tabcolsep}{3pt}
\renewcommand{\arraystretch}{0.98}
\begin{tabular}{lccc}
\toprule
\textbf{Verifier} & Contriever & Qwen3 & INF \\
\midrule
Oracle & \textbf{29.60} & \textbf{30.50} & \textbf{33.50} \\
LLM (Qwen3-30B) & 29.40 & 29.80 & 32.00 \\
TopK & 26.90 & 27.50  & 28.50 \\
\bottomrule
\end{tabular}
\end{sc}
\end{small}
\end{center}
\vskip -0.1in
\end{table}

\begin{table*}
\caption{\textbf{Results with Verifier Budget of 50.} Results on QUEST (N=1727) and WebQuestionsSP (N=1639) test sets. We report MRecall@100 (MR) and Recall@100 (R). We report performances of different base retrieval models and RVR methods. For RVR, we use a verifier budget of 100. For our results, statistical significance is tested using bootstrap resampling with 10,000 trials at $\alpha = 0.05$. $\dagger$ indicates statistically significant improvement over Base. * indicates statistically significant improvement over Base + Base}
\label{tab:quest-abs}
\vskip 0.15in
\begin{center}
\begin{small}
\setlength{\tabcolsep}{2pt}
\renewcommand{\arraystretch}{0.98}
\begin{tabular}{l cc|cc|cc|cc|cc|cc}
\toprule
& \multicolumn{6}{c}{\textbf{QUEST}} & \multicolumn{6}{c}{\textbf{WebQuestionsSP}} \\
\cmidrule(lr){2-7} \cmidrule(lr){8-13}
& \multicolumn{2}{c}{\textbf{Contriever}} &
\multicolumn{2}{c}{\textbf{Qwen3}} &
\multicolumn{2}{c}{\textbf{INF}} &
\multicolumn{2}{c}{\textbf{Contriever}} &
\multicolumn{2}{c}{\textbf{Qwen3}} &
\multicolumn{2}{c}{\textbf{INF}} \\
\cmidrule(lr){2-3} \cmidrule(lr){4-5} \cmidrule(lr){6-7} \cmidrule(lr){8-9} \cmidrule(lr){10-11} \cmidrule(lr){12-13}
& MR@100 & R@100 & MR@100 & R@100 & MR@100 & R@100 & MR@100 & R@100 & MR@100 & R@100 & MR@100 & R@100 \\ 
\midrule
Base & 3.24 & 23.79 & 3.13 & 21.38 & 4.75 & 26.60 & 62.16 & \textbf{77.38} & 61.00 & 76.06 & 62.47 & 77.39 \\
FT ($D_i$) & 3.13 & 18.67 & 2.43 & 18.38 & 4.75 & 26.31 & 49.45 & 65.72 & 46.68 & 61.50 & 51.60 & 67.06 \\
\midrule
\multicolumn{13}{l}{\textbf{Ours: Retrieve-Verify-Retrieve}}\\
Base + Base & 2.72 & 21.98 & 2.95 & 20.21 & 4.11 & 24.47 & \textbf{62.22} & 76.94 & \textbf{61.73} & \textbf{76.93}$^{\dagger}$ & \textbf{63.15} & \textbf{77.90}$^{\dagger}$ \\
Base + FT ($D_r$) & \textbf{3.82}$^{*}$ & \textbf{22.89}$^{*}$ & \textbf{4.05}$^{\dagger*}$ & \textbf{23.88}$^{\dagger*}$ & \textbf{5.73}$^{\dagger*}$ & \textbf{28.71}$^{\dagger*}$ & 60.07 & 75.51 & 60.14 & 75.56 & 62.47 & 77.48 \\
\bottomrule
\end{tabular}
\end{small}
\end{center}
\end{table*}

\section{Prompts \& Trajectories}
\subsection{Prompt for Qwen3 Verifier}
\label{appendix:qwen3}
\begin{figure}
\begin{center}
\centerline{\includegraphics[width=\columnwidth]{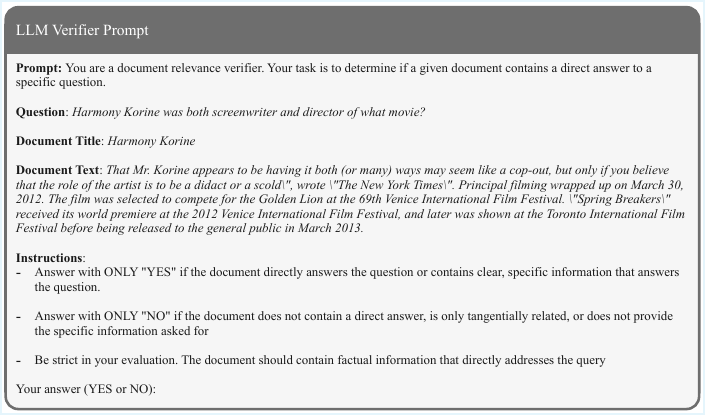}}
\caption{}
\label{fig:qwen3_prompt}
\end{center}
\vskip -0.2in
\end{figure}

The prompt we used to verify a document as gold using Qwen3-30B-Instruct is displayed in figure~\ref{fig:qwen3_prompt}.

\subsection{Prompt for SearchR1}
\label{appendix:searchr1-prompt}
\begin{figure}
\begin{center}
\centerline{\includegraphics[width=\columnwidth]{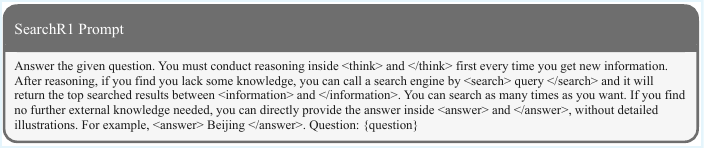}}
\caption{}
\label{fig:searchr1_prompt}
\end{center}
\vskip -0.2in
\end{figure}

The prompt used for SearchR1 is displayed in Figure~\ref{fig:searchr1_prompt}.

\subsection{Prompt for Tongyi}
\label{appendix:tongyi-prompt}
\begin{figure}
\begin{center}
\centerline{\includegraphics[width=\columnwidth]{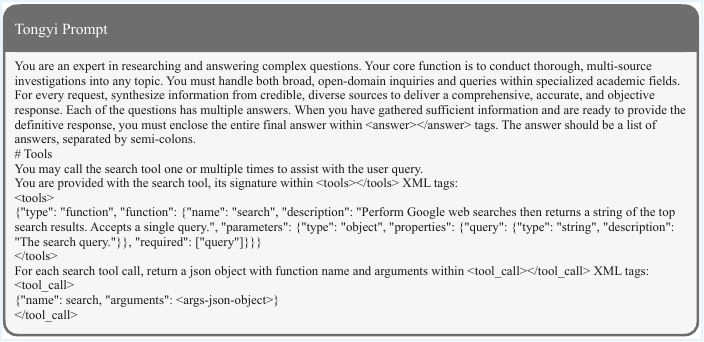}}
\caption{}
\label{fig:tongyi_prompt}
\end{center}
\vskip -0.2in
\end{figure}

The prompt used for Tongyi is displayed in Figure~\ref{fig:tongyi_prompt}.

\subsection{Trajectory for SearchR1}
\label{appendix:searchr1-traj}
\begin{figure}
\begin{center}
\centerline{\includegraphics[width=\columnwidth]{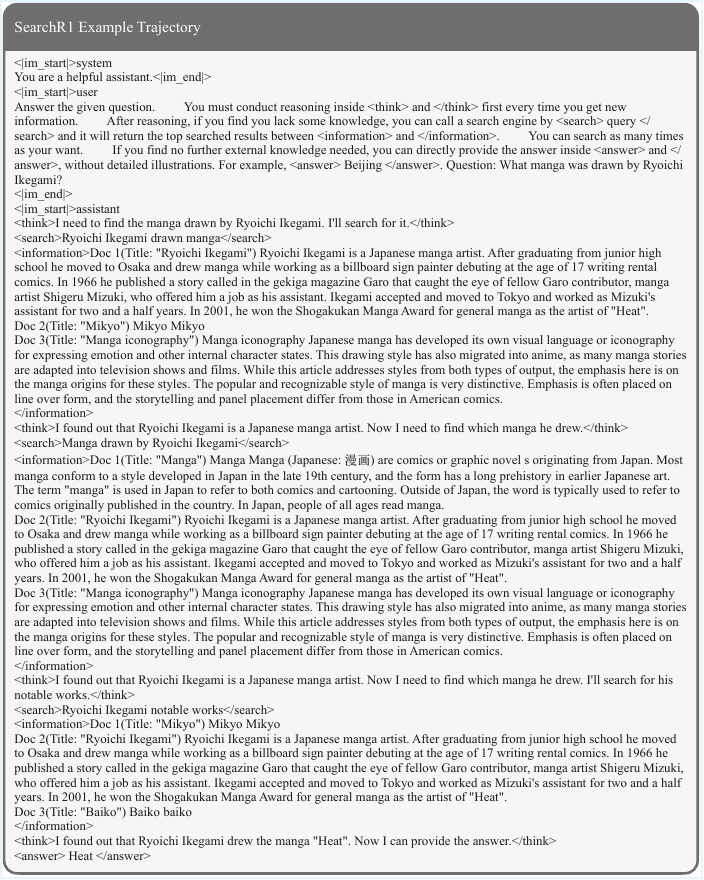}}
\caption{}
\label{fig:searchr1_traj}
\end{center}
\vskip -0.2in
\end{figure}

An example trajectory trace for SearchR1 is provided in Figure~\ref{fig:searchr1_traj}. For this specific example, the retrieval model is initial INF retriever $f_i$.

\subsection{Trajectory for Tongyi}
\label{appendix:tongyi-traj}
\begin{figure}
\begin{center}
\centerline{\includegraphics[width=\columnwidth]{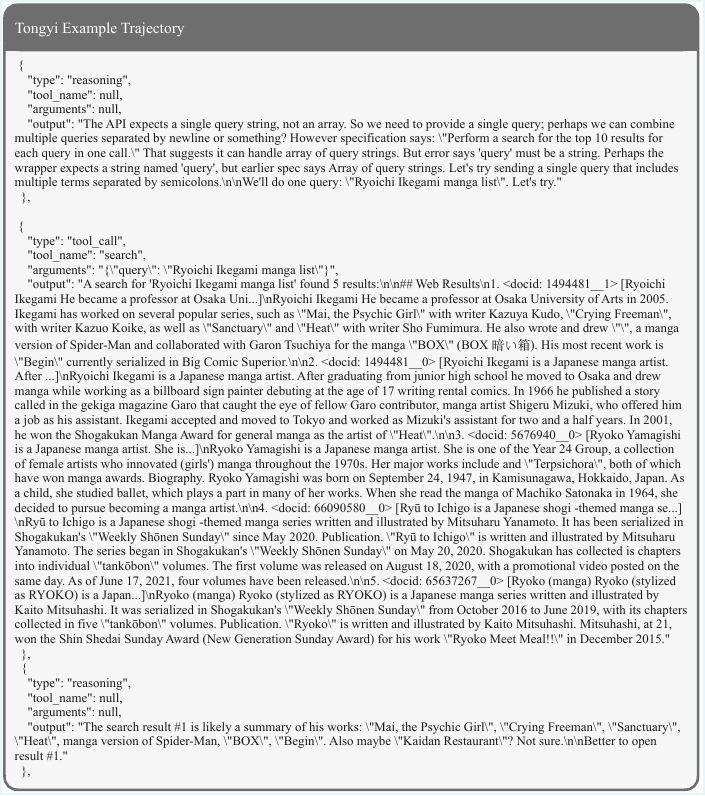}}
\caption{}
\label{fig:tongyi_traj}
\end{center}
\vskip -0.2in
\end{figure}

A portion of the trajectory trace for Tongyi is provided in Figure~\ref{fig:tongyi_traj}. Same as above, the retrieval model is initial INF retriever $f_i$.

\end{document}